\documentclass[arxiv]{ieeeaccess}
\usepackage{cite}
\usepackage{amsmath,amssymb,amsfonts}
\usepackage{algorithmic}
\usepackage{graphicx}
\usepackage{textcomp}
\usepackage[hidelinks]{hyperref}

\usepackage{pifont}
\usepackage[dvipsnames]{xcolor}
\newcommand{\cmark}{\color{Green}\ding{51}}%
\newcommand{\xmark}{\color{red}\ding{55}}%
\newcommand{\changed}[1]{\textcolor{black}{#1}}

\def\BibTeX{{\rm B\kern-.05em{\sc i\kern-.025em b}\kern-.08em
    T\kern-.1667em\lower.7ex\hbox{E}\kern-.125emX}}
\begin{document}

\ifarxiv\else
\history{Date of publication xxxx 00, 0000, date of current version xxxx 00, 0000.}
\fi
\doi{TBA}

\title{NICOL: A Neuro-inspired Collaborative Semi-humanoid Robot that Bridges Social Interaction and Reliable Manipulation}


\renewcommand{\thevol}{\textasteriskcentered}
\renewcommand{\theyear}{2023}

\author{\uppercase{Matthias Kerzel}\authorrefmark{1},
\uppercase{Philipp Allgeuer}\authorrefmark{1},
\uppercase{Erik Strahl}\authorrefmark{1},
\uppercase{Nicolas Frick}\authorrefmark{1},
\uppercase{Jan-Gerrit Habekost}\authorrefmark{1},
\uppercase{Manfred Eppe}\authorrefmark{2},
and \uppercase{Stefan Wermter}.\authorrefmark{1}}
\address[1]{Knowledge Technology, University of Hamburg, Hamburg, Germany (e-mail: \{matthias.kerzel,philipp.allgeuer,erik.strahl,stefan.wermter\}@uni-hamburg.de)}
\address[2]{Data Science Foundations Institute, Hamburg University of Technology, Hamburg, Germany (e-mail: manfred.eppe@tuhh.de)}

\tfootnote{The authors gratefully acknowledge support from the DFG (CML, MoReSpace, LeCAREbot), BMWK (VERIKAS), and the EU Commission (TRAIL, TERAIS).\\
The authors thank Phuong Nguyen for his contributions to the NICOL robot, and Sergio Lanza for his feedback on the manuscript.}

\markboth
{Kerzel \headeretal: NICOL: A Neuro-inspired Collaborative Semi-humanoid Robot}
{Kerzel \headeretal: NICOL: A Neuro-inspired Collaborative Semi-humanoid Robot}

\corresp{Corresponding author: Philipp Allgeuer (e-mail: philipp.allgeuer@uni-hamburg.de).}

\begin{abstract}
\changed{Robotic platforms that can efficiently collaborate with humans in physical tasks constitute a major goal in robotics. However, many existing robotic platforms are either designed for social interaction or industrial object manipulation tasks. The design of collaborative robots seldom emphasizes both their social interaction and physical collaboration abilities.
To bridge this gap, we present the novel semi-humanoid NICOL, the \emph{Neuro-Inspired COLlaborator}. NICOL is a large, newly designed, scaled-up version of its well-evaluated predecessor, the Neuro-Inspired COmpanion (NICO). NICOL adopts NICO's head and facial expression display and extends its manipulation abilities in terms of precision, object size, and workspace size.
Our contribution in this paper is twofold---firstly, we introduce the design concept for NICOL, and secondly, we provide an evaluation of NICOL's manipulation abilities by presenting a novel extension for an end-to-end hybrid neuro-genetic visuomotor learning approach adapted to NICOL's more complex kinematics. We show that the approach outperforms the state-of-the-art Inverse Kinematics (IK) solvers KDL, TRACK-IK and BIO-IK.
Overall, this article presents for the first time the humanoid robot NICOL, and contributes to the integration of social robotics and neural visuomotor learning for humanoid robots.}

\end{abstract}


\ifarxiv
\titlepgskip=-21pt
\else
\titlepgskip=-13pt
\fi

\maketitle

\section{Introduction}

What are the main prerequisites for a collaborative robot to act and work alongside humans? On the one hand, the robot needs to be able to perform precise and reliable object manipulation in a large workspace, but on the other hand, the robot also needs to be able to interact with human coworkers on a social level. This requires auditory communication as well as the understanding and synthesis of both social cues and non-verbal communication like gestures and facial expressions.

\begin{figure}
    \centering
    \includegraphics[width=1.0\columnwidth]{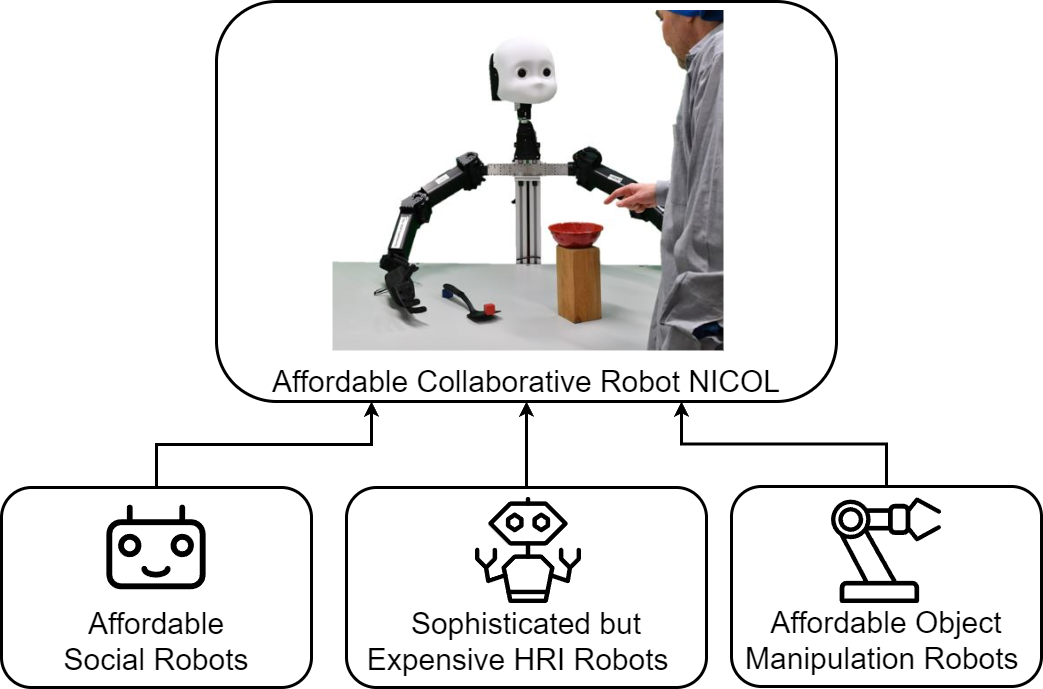}
    \caption{NICOL is an affordable robot  platform that bridges the gap between social robotics and reliable object manipulation in a large workspace.}
    \label{fig:nicol_bridging_gap}

\end{figure}

Social interaction is one of the crucial factors for intuitive human-robot collaboration and the ability of robots to learn from humans \cite{Thomaz:2006, Srinivasan:2016, KPSHW20}. Gaze, as a non-verbal social cue, facilitates shared attention, which can signal future actions \changed{or draw attention toward} an object or location \cite{mutlu2013coordination}. Likewise, facial expressions can be used to give feedback or warn about dangers. Speech output also offers an intuitive way to request aid and instructions, or to suggest a course of action. There exist only very few robots, however, that integrate social interaction with reliable manipulation of adult-scale objects while remaining affordable. A robot is desirable that bridges this gap (see Fig.~\ref{fig:nicol_bridging_gap}). \changed{See Section \ref{related_platofrms} for an overview of related platforms.}



The social interaction capabilities of NICOL are based on the well-evaluated design of its predecessor, the humanoid child-sized Neuro-Inspired COmpanion (NICO) \cite{KSMNHW17}. Like NICO, NICOL has an articulated head that can express stylized facial expressions and focus its gaze on interaction partners or relevant elements of the shared workspace. While NICO has limited low-precision object manipulation abilities, NICOL can manipulate objects \mbox{using} two Robotis OpenManipulator-P arms with adult-sized five-fingered Seed Robotics RH8D hands attached. It has a workspace similar to an adult person and can handle everyday objects.

In addition to the hardware, NICOL also features a software framework that is directed towards two groups of users---the roboticists that wish to work at a low level with a high degree of control, and the users from different backgrounds, like artificial intelligence, machine learning, cognitive modelling and human-robot interaction, that require a simpler and more abstract interface. Low-level control is delivered via the Robot Operating System (ROS) and provides a rich control interface with great customizability and extensibility. Manipulating and developing within a ROS workspace often requires many non-trivial software cross-dependencies and requires a certain level of expertise to navigate, use and maintain. For users with little background in robotics, a more lightweight system is desired. Furthermore, such users often demand a lightweight simulated version of the robot \changed{for performing rapid iterations of parallelized deep reinforcement learning experiments.}
NICOL meets the demands of both groups---roboticists and AI researchers. Its software offers a full ROS interface, but users can also connect to a front-end Python interface running on top of ROS, which enables them to directly use the robot without having to explicitly deal with ROS. In addition, we provide a simulated virtual version of NICOL.

\changed{We introduce NICOL as a novel robot platform; we provide a thorough evaluation of the main advances of NICOL over its predecessor, NICO, its enhanced manipulation abilities and a novel extension to a visuomotor learning approach. In detail, our contributions are:
\begin{itemize}
    \item We place NICOL in context by examining existing related platforms (Section \ref{sec:rel_work}). This demonstrates how our robot fills the gap between social robotics and object manipulation in an affordable manner.
    \item Section \ref{sec:nicol} presents the hardware and software concept for NICOL, the \emph{Neuro-Inspired COLlaborator}. Herein, we also introduce the social interaction capabilities of NICOL, including face and emotion generation capabilities as well as auditory and visual capabilities.
    \item Section \ref{sec:vis_method} presents a thorough evaluation of NICOL's kinematics. We adapt and extend different neural and hybrid neuro-genetic visuomotor approaches initially developed for NICO to the larger NICOL and its more complex kinematics.
    We show the challenges of neural end-to-end learning for complex kinematics and successfully evaluate a hybrid neuro-genetic approach.
    \item Section \ref{sec:ik_vis_learning} introduces a novel neuro-genetic visuomotor learning approach that improves the learned grasping accuracy of NICOL to over 99\%, and contributes to the growing research on neuro-robotic approaches that are suitable for transfer between platforms with different morphologies \cite{mivseikis2018transfer}.
    Here, we also present experiments  to show that the grasp accuracy of NICOL improves to over 99\%, outperforming the state-of-the-art inverse kinematics (IK) solvers KDL, TRACK-IK and BIO-IK. 
\end{itemize}
}
Overall, this article presents for the first time NICOL, which serves as an affordable research platform that integrates social robotics and humanoid manipulation, and as a testbed for neural visuomotor learning.

\begin{table*}
\setlength{\tabcolsep}{3pt}
\centering
\begin{tabular}{|l|c|c|c|c|c|c|c|c|c|c|}
\hline
Robot & Type & Humanoid Body & Humanoid Head & Facial Expressions & Gestures & Grasping & Workspace & Price Range\\ 
\hline
UR-5                & robot arm  & \xmark & \xmark & \xmark &\xmark & \cmark & adult-sized & mid \\
KUKA-DLR Arm        & robot arm  & \xmark & \xmark & \xmark &\xmark & \cmark & adult-sized & mid \\
Franka Research 3   & robot arm  & \xmark & \xmark & \xmark &\xmark & \cmark & adult-sized &  low \\
\hline
Furhat              & robot head & \xmark & \cmark & \cmark & \xmark & \xmark & none & mid \\
iCub head           & robot head & \xmark & \cmark & \cmark & \xmark & \xmark & none & mid \\
\hline
ROBOTIS OP3         & infant-sized robot & \cmark & \xmark & \xmark & \cmark & \xmark & tiny & low \\
Nao                 & infant-sized robot & \cmark & \cmark & \xmark & \cmark & \cmark & tiny & low \\
\hline
iCub                & child-sized robot & \cmark & \cmark & \cmark & \cmark & \cmark & child-sized & high \\
NimbRo-OP           & child-sized robot & \cmark & \cmark & \xmark & \xmark & \xmark & child-sized & mid \\
Pepper              & child-sized robot & (\cmark) upper body & \cmark & \xmark & \cmark & \xmark & child-sized & mid \\
Poppy         & child-sized robot & \cmark & \cmark & \cmark & \cmark & \cmark & child-sized & low \\
NICO (ours)         & child-sized robot & \cmark & \cmark & \cmark & \cmark & \cmark & child-sized & low \\
\hline
Talos               & adult-sized robot & \cmark & \cmark              & \xmark & \cmark & \cmark & adult-sized & high \\
PR2                 & adult-sized robot & \xmark & \xmark              & \xmark & \xmark & \cmark & adult-sized & high \\
Sawyer              & adult-sized robot & \xmark & \xmark              & \cmark & \xmark & \cmark & adult-sized & mid \\
ARMAR-6             & adult-sized robot & (\cmark) upper body & \xmark & \xmark & \cmark & \cmark & adult-sized & high \\
TOMM                & adult-sized robot & (\cmark) upper body & \xmark & \xmark & \xmark & \cmark & adult-sized & high \\
Fetch               & adult-sized robot & \xmark & \xmark & \xmark & \xmark & \cmark & adult-sized & high \\
R1                  & adult-sized robot & (\cmark) upper body & \xmark & \cmark & \cmark & \cmark & adult-sized & high \\
Sophia              & adult-sized robot & (\cmark) upper body & \cmark & \cmark & \cmark & \xmark & adult-sized & high \\
NICOL (ours)        & adult-sized robot & (\cmark) upper body & \cmark & \cmark & \cmark & \cmark & adult-sized & mid \\
\hline
\end{tabular}
\caption{Robotic platforms for social interaction and object manipulation. Price ranges are estimated; a low price range is below 10,000 EUR, a medium price range is below 100,000 EUR and a high price range is above this. Few platforms are designed for both social interaction and physical tasks while still being affordable. NICOL fills the gap of a robotic platform with both an upper humanoid body and humanoid head, capable of facial expressions, gestures and object manipulation in an adult-sized workspace at an affordable price tag.}
\label{table1}
\end{table*}

\section{Related Work}
\label{sec:rel_work}

\subsection{Related Robotic Platforms}
\label{related_platofrms}
NICOL combines the capabilities of dexterous robotic manipulation and social interaction, and so has different types of robots as related platforms.

There are many robotic manipulators for industry and research. Some common platforms are the UR-5\cite{UR5} and its related designs from Universal Robots, as well as the KUKA-DLR Lightweight Robot arm\cite{bischoff2010kuka}, and the Franka Research 3\cite{pandarobot}. These robotic arms have at least six degrees of freedom (DoF) to ensure solvable inverse kinematics for general 6-dimensional poses.
These manipulators and their various end-effectors are primarily designed to handle objects and lack the social interaction capabilities that enable intuitive learning from, or teaching by, humans.

On the other side of the spectrum are platforms designed for social interaction with no, or limited, manipulation capabilities. For instance, the iCub \cite{metta2010icub} is available as a stand-alone 3-DoF head that can display facial expressions and perform gaze shifts. The Furhat \cite{al2012furhat} robot head is another example and can project animated or recorded faces for interaction purposes.

Infant-sized or toy-sized humanoids form another type of social platform. Their small size is advantageous for research, they have an affordable price, and they are inherently safe due to low motor speeds and weight. Well-known examples are the Softbank Robotics NAO (formerly Aldebaran) and the ROBOTIS OP3 (descendant of the DARwIn-OP)\cite{ha2011development}. While infant-sized platforms often have humanoid manipulators, they are too small to manipulate adult-sized items or reach them in domestic-scale environments.

The next larger category of humanoids is child-sized. A popular platform for developmental research is the iCub\cite{metta2010icub}. Its size resembles a child of about 90 cm, it is well-actuated with 53 DoF, features human-like hands and gaze shifts, and can display stylized facial expressions. NICOL is significantly more affordable and simple than the iCub, and is thereby less challenging to modify and/or customize for particular needs.
The NimbRo-OP family of robots \cite{schwarz2013humanoid, AllgeuerIguhop, FichtNop2X} are designed for the RoboCup soccer league, and focus on bipedal walking with arms for balance as well as getting up, but not on manipulation. As such, they do not even have hands that can be actuated.
Softbank's Pepper robot \cite{Pandey:2018:pepper} features a humanoid torso on top of an omnidirectional wheeled platform. It has 20 DoF in the upper body, but its arms and hands are designed for gesturing, not grasping. 
Poppy is an open-source 3D-printed robot designed for education, artists, and scientists \cite{Lapeyre2014_Poppy}. It is completely customizable and comes with a display to show facial expressions and other information. 
Poppy can perform bipedal locomotion, but in its primarily advertised configuration, it features only very limited grasping functionalities. 
The concept of NICO is quite similar to that of Poppy. NICO is an open platform for researching neuro-robotic models for human-robot interaction, as well as visuomotor learning \cite{KSMNHW17}. Its head adapts the open iCub design, and, in contrast to Poppy, integrated LED arrays display stylized facial expressions. Furthermore, NICO has quite sophisticated grasping abilities to manipulate small objects, provided by two 6-DoF arms with fully functional child-sized anthropomorphic hands produced by Seed Robotics.

Adult-sized humanoids with manipulation abilities can generally handle real-life objects. However, such platforms are often expensive and difficult to maintain---like PAL's Talos\cite{stasse2017talos}, which was introduced at a price of about 1 million euros---or their design is too non-humanoid for social interaction, like the Atlas initially developed by Boston Dynamics, and the PR2 from Willow Garage. Some adult-sized platforms lack object manipulation abilities entirely, like the Hanson Robotics Sophia robot. The ARMAR-6 from the KIT \cite{asfour2019armar} is designed for complex collaborative tasks in industrial environments, but it neither features a human-like face nor can it display facial expressions. Likewise, the TOMM \cite{dean2017tomm} and the one-armed Fetch robot \cite{fetch} focus on manipulation and wheeled mobility, but not on social cues. The R1 from the IIT \cite{parmiggiani2017designR1} is a wheeled platform with grippers that can display an animated face on a screen. A related design is the Rethink Robotics Sawyer, the successor of the well-known humanoid Baxter robot. It features a tablet for displaying animated eyes, but it only has a single end effector and is not humanoid.

Table~\ref{table1} summarizes our analysis of robotic platforms. It is evident that there is a gap in available platforms that are designed for both social interaction and adult-sized object manipulation tasks. \changed{The few platforms that fulfill} these two criteria are prohibitively expensive. We propose the open NICOL design to address this gap in the state of the art.
In this article, we show that the NICOL is capable of social interaction tasks by extrapolating from previous work on the NICO, its ``smaller brother''. NICO and NICOL share the design of their head and facial expression mechanism(s), but importantly, the latter possesses a significantly increased arm strength and workspace size, allowing for more real object scenarios.

\begin{figure*}[ht]
    \centering
    \includegraphics[width=1\linewidth]{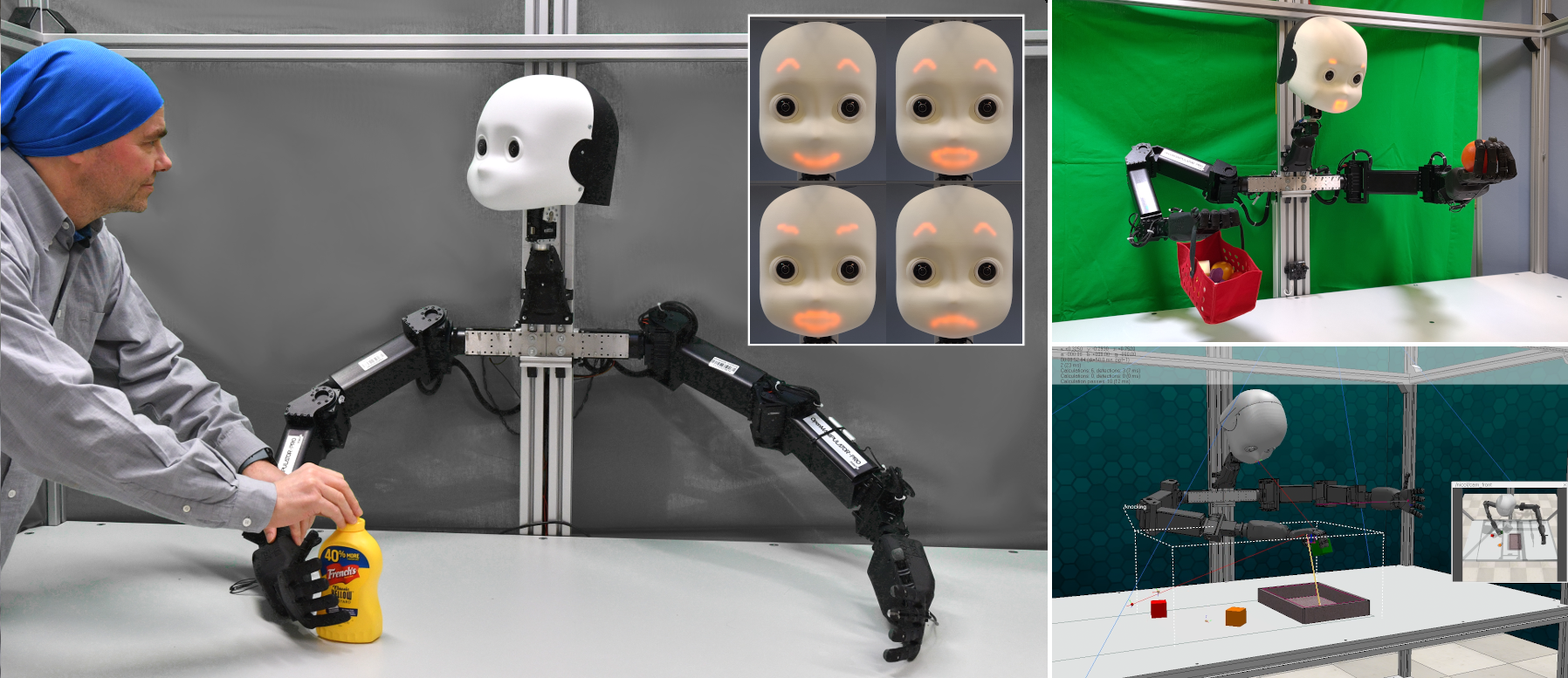}
    \caption{NICOL semi-humanoid platform. Left: NICOL jointly works on a grasping task with an experimenter (in the upper-right corner thereof, different facial expressions of NICOL are shown that can give feedback during the task). Right: NICOL object manipulation and expression in real life (top) and simulation (bottom).}
    \label{fig:NICOL}
\end{figure*}

\section{Semi-humanoid Robotic Platform NICOL}
\label{sec:nicol}
NICOL is designed for research on robots that learn from and collaborate with humans. It can serve as a platform for integrating neuro-robotic visuomotor models combined with affective interaction, joint attention, and shared perception. NICOL, depicted in Fig.~\ref{fig:NICOL}, consists of three main components: two manipulators with anthropomorphic hands, and the head with a facial expression display.
NICOL adapts the head design of the child-sized humanoid NICO and combines it with two Robotis manipulators with five-fingered hands. NICOL can naturally collaborate with human partners on physical tasks, thanks to its ability to manipulate adult-size domestic objects and to use non-verbal cues like gaze or facial expression.

\subsection{Social Interaction Capabilities of NICOL}
The interaction capabilities of NICOL are based on the design of NICO. Here, we summarize the main findings and studies.
The seven facial expressions of NICO, which are now shared by NICOL, are evaluated by Churamani et al.\ \cite{CKSBW17}. In a study with twenty participants from eleven different countries, five expressions (neutral, happiness, sadness, surprise and anger) are recognized by the participants with an accuracy of over 75\%. The positive effect of the emotion display on the robot's subjective user rating is verified with a Godspeed questionnaire \cite{Bartneck:2009}. The freely programmable LED arrays are also used in learning emotion expression via reinforcement learning \cite{CBSW18}.
Ng et al. \cite{NABCFHMMNNSSGHNSTWW17} use facial expressions in combination with neurocognitive models for social cue recognition and behavior control, and report that a natural language dialogue system benefits from the robot directly looking at the face of its interaction partner. Together with a more personalized conversation, this behavior made participants perceive NICO as more \changed{intelligent and likable}. Beik-Mohammadi et al. \cite{BXABCNNSAGHSWW19} show that using social gestures and more socially engaging dialogue enhanced the robot's perceived likeability and animacy. Finally, Kerzel et al. \cite{KPSHW20} show that participants guided by NICO in a visuomotor learning scenario via verbal requests and facial expressions, as compared to instructions by a human experimenter, rated the human-robot interaction as more positive and engaging. This increased engagement can improve the outcome of the learning scenario.

In summary, the studies strongly suggest that the design of NICOL's head, which is adapted from NICO, can effectively create social cues in terms of facial expressions and gaze. They also show that these social cues positively affect human-robot interactions. 

\subsection{Arms and Hands}

\begin{figure*}[ht]
    \centering
    \includegraphics[width=1\linewidth]{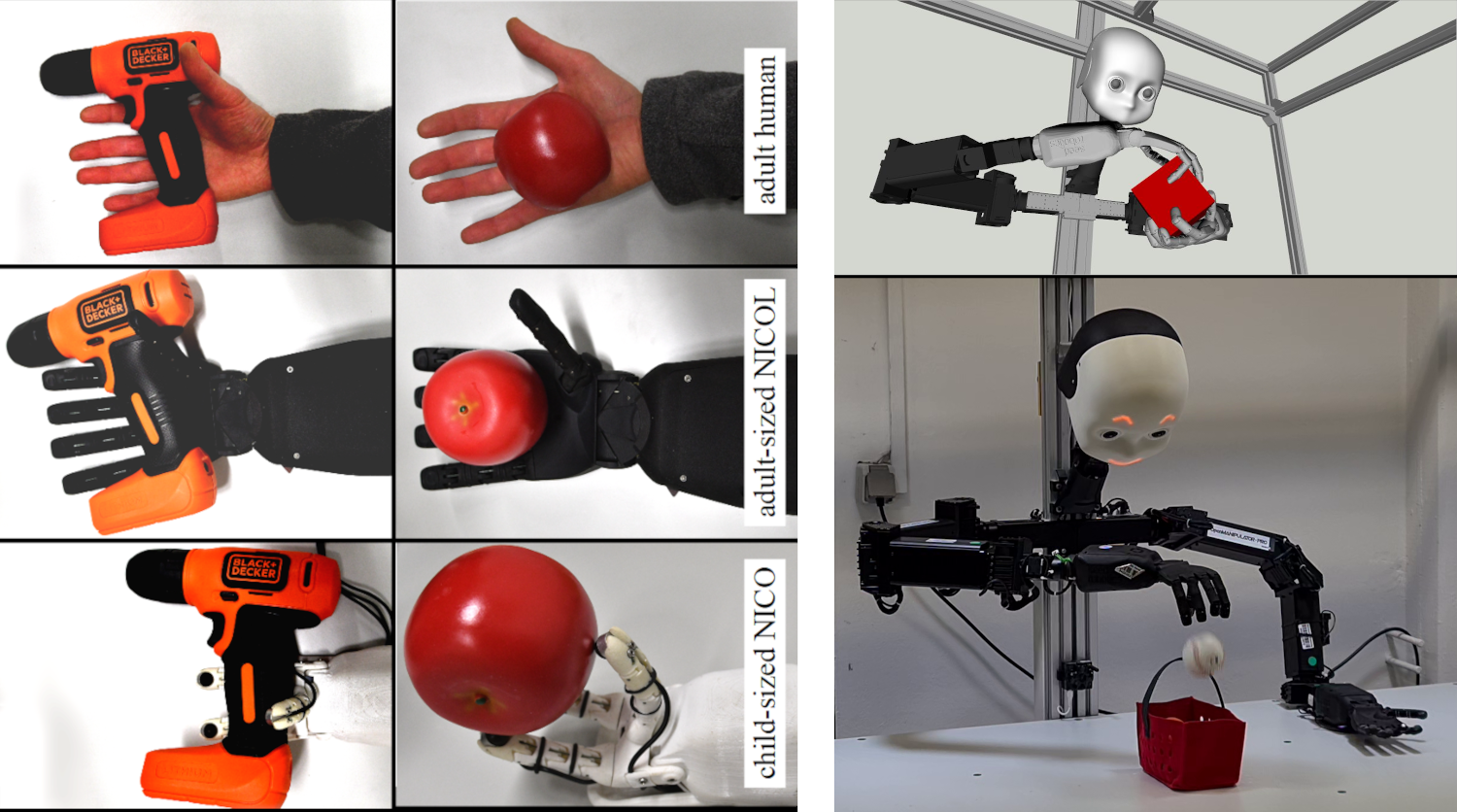}
    \caption{Left: Size comparison between an adult human's hand, NICOL, and NICO. Upscaling the hands of NICO allows NICOL to manipulate everyday objects and tools. Right: NICOL in a virtual Gazebo environment during a dexterous manipulation task and dropping a baseball into a basket.}
    \label{fig:gazebo_new}
\end{figure*}

The arms of NICOL consist of two Robotis OpenManipulator\nobreakdash-P\footnote{https://www.robotis.us/openmanipulator-p} arms with 6 DoF and a payload of 3\,kg. As end-effectors, two SeedRobotics RH8D adult-sized robot hands\footnote{https://www.seedrobotics.com/rh8d-adult-robot-hand} with a manipulation payload of 750\,g are used. All five fingers in the hand are tendon-operated. Each three-segment finger is operated by a single tendon. Each hand has eight actuated DoF---three in the wrist for rotation\footnote{Note that the wrist rotation is redundant and therefore not used.}, flexion, and abduction, two DoF in the thumb for abduction and flexion, and one DoF each for index finger flexion, middle finger flexion and combined flexion of the ring and little finger. The tendon mechanism allows the hand to coil around objects without further fine control.
The arms of NICOL can reach up to 100 cm and therefore have a workspace slightly larger than that of an adult sitting at a table.

\subsection{Head and Audiovisual Sensing}
The design of the shell of NICO's head is based on a modified version of the open-source iCub design\cite{metta2010icub}. The design balances human features with enough abstraction to avoid the uncanny valley effect\cite{mori2012uncanny}. Stylized facial expressions can be displayed by NICOL using three LED arrays that are placed behind the eyes (two 8x8 LED arrays) and mouth (one 16x8 array). Due to the head's material properties, the individual LEDs form lines shining through the head's material, as shown in Fig. \ref{fig:NICOL}.
An internal speaker is used to facilitate spoken communication. The head is articulated with two DoFs for pitch and yaw movements. Two See3CAM CU135 cameras with a  4096 x 2160 (4K) resolution are mounted in the eye sockets of NICOL. Their fisheye lens has a field of view of 202 degrees.
Two Soundman OKM II binaural microphones are placed at the sides of NICOL's head; due to the absence of head-internal fans or mechanics, the robot's ego noise is very low.

\subsection{Table Environment and Safety Measures}
The table environment measures 100 x 200 cm and is located at a height of 74 cm. The head and arms of the NICOL are mounted to a vertical support at the \changed{rear center of the table.} Scaffolding at the corners of the table creates a visible delimitation for the workspace of the robot.
A human interaction or collaboration partner is thereby implicitly made aware when actively reaching into the robot's workspace. For additional safety, two emergency shutdown buttons are integrated within the operator's workspace. The scaffolding can further be used to mount external sensors or devices, like cameras or light sources.

\begin{figure*}[ht]
    \centering
    \includegraphics[width=1\linewidth]{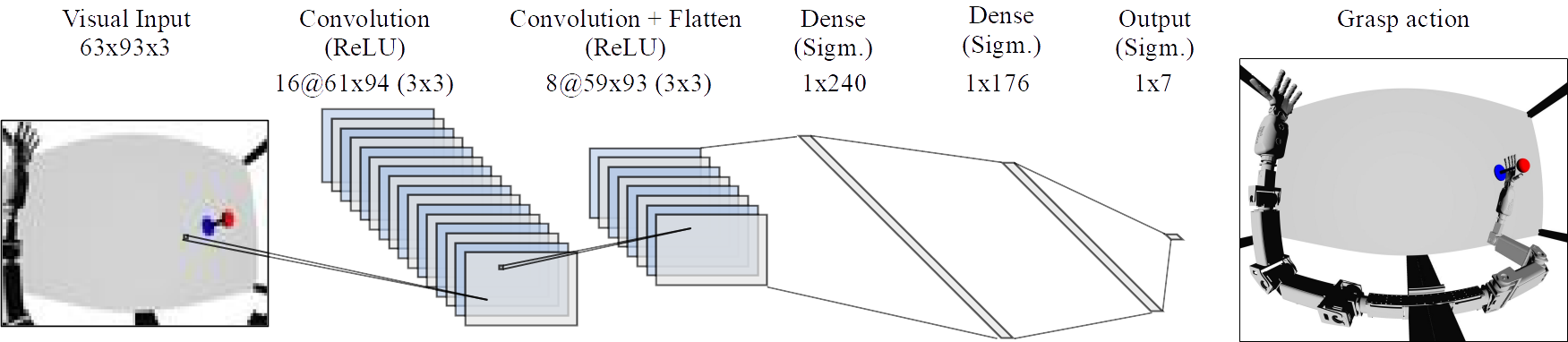}
    \caption{Neural architecture for end-to-end visuomotor learning.}
    \label{fig:neural_architecture}
\end{figure*}

\begin{figure*}[ht]
    \centering
    \includegraphics[width=1\linewidth]{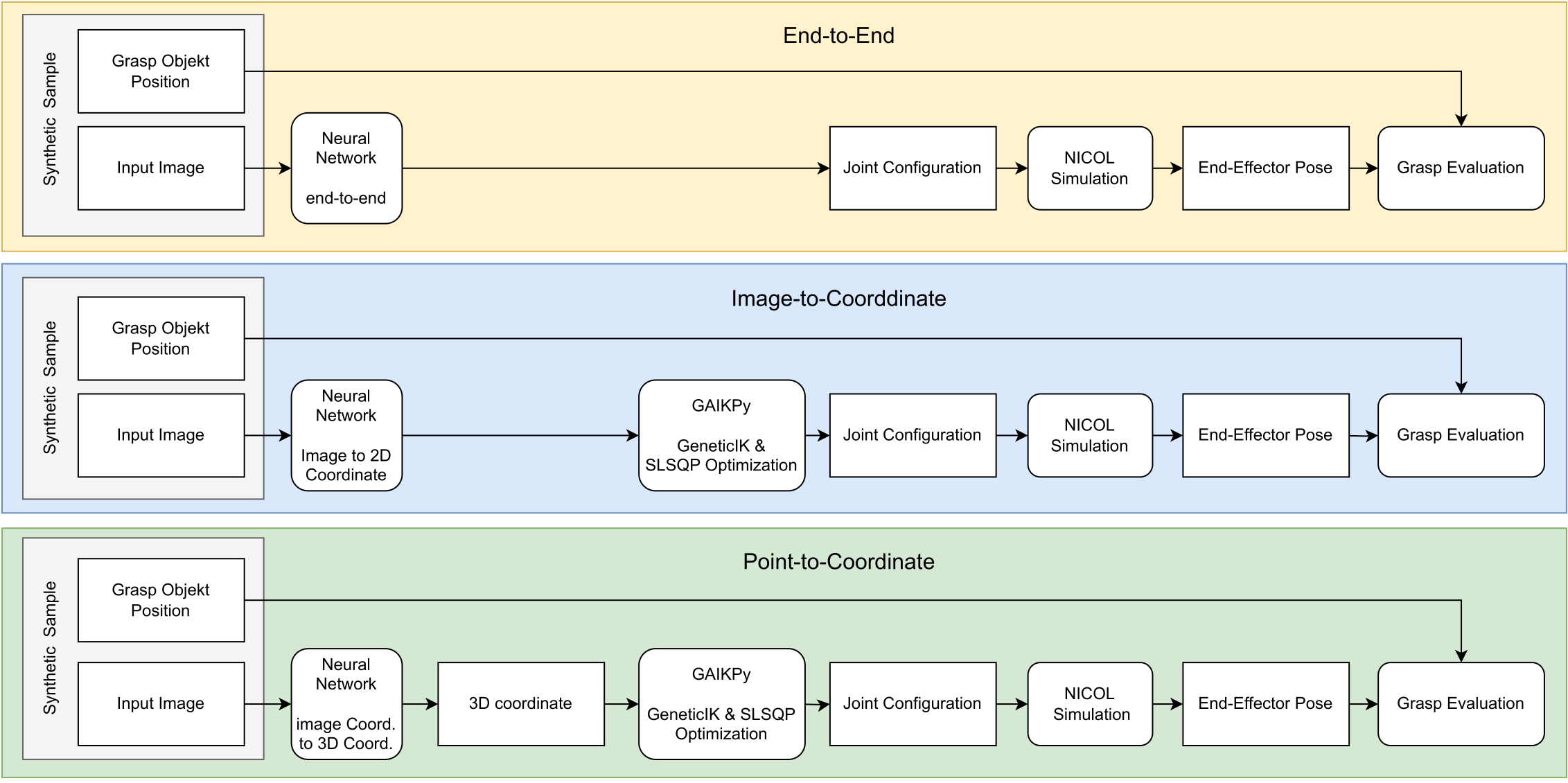}
    \caption{Three architectures for visuomotor learning.}
    \label{fig:CONDITION}
\end{figure*}

\subsection{Software and Simulation Environment}
The NICOL API is based on the Robot Operating System (ROS) middleware. All functionalities are provided through a modified version of the hardware controller delivered with the Robotis arms, extended to support the Seed Robotics hands, additional sensors, as well as drivers for custom hardware (facial expressions). The API integrates the MoveIt planning framework and provides a Python-based client. For prototyping and development, as well as for simulation of real-world scenarios, NICOL is realized in both the Gazebo simulation environment (see Fig.~\ref{fig:gazebo_new}) and CoppeliaSim (see Fig.~\ref{fig:NICOL}). The robot description is based on a URDF model, making it possible to import NICOL in many other simulators and environments.\footnote{Software and CAD/URDF files, including everything needed to run the CoppeliaSim simulation, are available at \url{https://www.inf.uni-hamburg.de/en/inst/ab/wtm/research/neurobotics/nicol.html}.} 

\section{Visuomotor Learning and State-of-the-Art IK Solvers}
\label{sec:vis_method}
To evaluate NICOL as a research platform for neurorobotic visuomotor learning, we first transfer an end-to-end neural visuomotor learning approach from NICO, a smaller humanoid robot, to NICOL. Based on the results, we propose two variations of a modular neuro-hybrid approach. In these modular approaches, we use a neural network for image processing and different \changed{IK solvers for the kinematics.} We evaluate the baseline performance of state-of-the-art IK solvers that are established in the robotics community and suggest a novel hybrid neuro-genetic approach that can handle the added kinematic complexity of a humanoid.

\subsection{State-of-the-Art IK Approaches}
Solving IK is a well-known and studied problem in the robotics community. The \textit{MoveIt planning framework} \cite{coleman2014reducing}, as part of the ROS middleware, offers the required flexibility to utilize different IK solvers for an experimental setup. It enables a plug-in-based configuration of inverse kinematics solvers as well as high-level motion planners. Mainly third-party kinematics libraries, contributed by foundations or private sector organizations are available, and these have become state-of-the-art due to the popularity of the ROS framework. The inverse kinematics and planning functionalities are provided in the form of ROS services and offer a variety of parameterized kinematics calculations.

The \textit{Kinematics and Dynamics Library} (KDL) by Orocos \cite{kdl-url} is de facto the most widely adapted inverse kinematics solution in the ROS ecosystem. It supports kinematic chains with a minimum of 6 DoF. KDL utilizes the well-known pseudo-inverse Jacobian method to determine suitable joint configurations. As the Jacobian matrix holds the partial derivatives between joint space and Cartesian space, the difference between the so-called seed state (typically the initial joint configuration) and the target state can be calculated in Cartesian space. Forward kinematics are used on the seed state to determine the current end-effector pose. A suitable joint configuration for achieving the target Cartesian end effector pose can then be obtained by calculating the error between the initial pose and the target pose, and multiplying it with the inverted Jacobian. The algorithm makes use of the Newton-Raphson method and iteratively minimizes the error between the previous and next target poses.

With the intention of improving the coverage of the robot workspace, TracLabs identified key issues in the KDL framework and proposed \textit{TRAC-IK} in 2015 \cite{tracikBeeson}. Remarkable about the approach is that two IK solvers are executed in parallel. One is a modified reimplementation of KDL called KDL-RR. The method additionally checks for local optima during the iterative improvement step and moves the next seed far enough away from them. The second solver, SQP-SS, formulates the IK problem as a non-linear sequential quadratic programming optimization problem. TRAC-IK returns the solution of the solver that first terminates with a valid joint configuration. In contrast to the pseudo-inverse Jacobian, SQP-SS is particularly capable of directly handling constraints, such as the joint limits, during the optimization step. This ability is particularly relevant  for humanoid robots like NICOL, whose joints are constrained to mimic the human range of motion.

Starke et al. introduced the \textit{Bio-IK} method \cite{bioikStarke} \cite{bioikmoveitRuppel}. The algorithm combines multiple bio-inspired optimization methods, most importantly evolutionary and particle swarm optimization, in order to solve the inverse kinematics problem. In difference to KDL and TRAC-IK, it does not rely on any Jacobian mathematics. The evolutionary algorithm is fundamental to the approach. Every individual in the population corresponds to a joint space robot pose. Momentums are assigned to every individual by hybrid particle swarm optimization, and these momentums are continuously updated during runtime. Besides selection, mutation and recombination, also elitism is used to prevent the deterioration of \changed{already-found solutions}. Local search is executed on the elites, and simulates mutations on single genes. The classical MoveIt interface only allows setting the pose or position goals. \textit{Bio-IK} additionally offers extended functionality in a separate service \cite{bioikserviceGithub}, where multiple custom goal types are available, e.g. linear end-effector trajectories or minimal joint displacement configurations. The various custom goal types can be combined into a single request, as each custom goal type is treated as a weighted partial cost function by the algorithm. The service also offers an approximate mode, leading to lower precision in the IK solutions but enabling higher coverage of the workspace. 

\subsection{End-to-end Neurorobotic Visuomotor Learning}
The supervised neural end-to-end visuomotor approach, first introduced in \cite{KW17}, learns to map a single object's visual input in the robot's workspace to joint configurations for reaching for the object. The approach was evaluated with a grasp success rate of approximately 85\% on NICO with 5 DoF in a workspace of size 30x40\,cm and a training set of 400 samples. In this paper, we evaluate if the larger workspace and more complex NICOL kinematics influence the learning outcome. The neural architecture is shown in Figure~\ref{fig:neural_architecture}. In Section~\ref{sec:vis_exp:neural_learning} we report experimental results.

\subsection{Hybrid neuro-genetic visuomotor learning}
Genetic algorithms are based on evolutionary selection, recombination and mutation processes of a population of individuals, modelled by their chromosomes \cite{marslandmachine}. Each chromosome encodes a potential solution to a task. Following Kerzel et al. \cite{KSSW20}, the chromosomes encode a joint configuration to reach a given pose. During each iteration, these chromosomes, expressed as individuals, are ranked according to their fitness, i.e., the position and orientation error compared to the goal pose. Successful individuals are copied into the next generation, with possible random mutations to their chromosome-encoded joint configuration. Niching is used to preserve diverse chromosomes; on every CPU of the computer system, a different evolution is implemented, \changed{thus fully utilizing multiple cores with minimal overhead.}

While genetic algorithms are adept at escaping local minima, they often have issues further optimizing found solutions. To overcome this limitation, we hybridize the genetic algorithm with gradient-based Sequential Least SQuares Programming (SLSQP). This computationally expensive procedure is only applied to the n-best individuals of the population. While preliminary experiments have indicated that SLSQP is prone to be attracted by local minima, the initialization with solutions found by the genetic algorithm can overcome this shortcoming. Optimized hyperparameters for the genetic algorithm and SLSQP were adapted from \cite{KSSW20}, where they yield good results for NICO. 

While the hybrid genetic algorithm provides IK solutions, a neural component is utilized to determine the position of the grasp object from visual input. \changed{The architecture is shown in Fig.~\ref{fig:CONDITION}, \changed{and experiments} are reported in Section \ref{sec:vis_exp:neural_learning}.}


\section{IK and Visuomotor Learning Experiments}
\label{sec:ik_vis_learning}
We first curate a new dataset (Section \ref{sec:vis_exp:dataset}) for visuomotor learning on the NICOL platform. A first analysis of the dataset allows us to get an insight into NICOL's workspace. In  Section \ref{sec:vis_exp:neural_learning}, we evaluate how well our neural end-to-end and a hybrid neuro-genetic visuomotor learning approach can be transferred from the child-sized humanoid NICO to the adult-sized NICOL. We evaluate how the significantly increased workspace and the more complex manipulator affect the learning outcome. \changed{In Sections \ref{sec:vis_exp:neurogenetic:color_based_loc} and \ref{sec:vis_exp:neurogenetic:image_to_table},} we present a modular neuro-hybrid approach in which we evaluate state-of-the-art IK solvers and a neuro-generic approach based on GAIKPy.

\subsection{Dataset for Visuomotor Learning Analysis of NICOL's Workspace}
\label{sec:vis_exp:dataset}
We collect three datasets\footnote{The datasets are available at \url{https://www.inf.uni-hamburg.de/en/inst/ab/wtm/research/corpora.html}.}, each consisting of 10,000 samples in a virtual environment. The datasets differ in the number of active joints of NICOL, ranging from six to eight. As the joints of NICOL are constrained to mimic the human range of motion, we hypothesize that the additional degrees of freedom beyond six will increase the number of reachable grasp poses.
Each sample contains the following: 1) an image from the egocentric perspective of NICOL with a grasp object placed onto the table on the right side of the workspace in a 100 x 100 cm area; the images are cropped and resized to 63 x 96 pixels. 2) The x- and y-coordinates of the grasp object on the table. 3) The configuration of NICOL's 8 arm joints that result in a grasp pose. The grasp pose has the hand pointing forward with the palm touching the grasp object at the middle of its height, as shown in Fig. \ref{fig:neural_architecture} (right side). We use a genetic algorithm \emph{GAIKPy} introduced in \cite{KSSW20} for computing the joint configuration for grasping the object that is placed at a random position within the robot's workspace. If no suitable joint configuration can be found with the genetic algorithm, the sample is rejected, and another random object position is generated. 

\begin{figure*}[ht]
    \centering
    \includegraphics[width=01\linewidth]{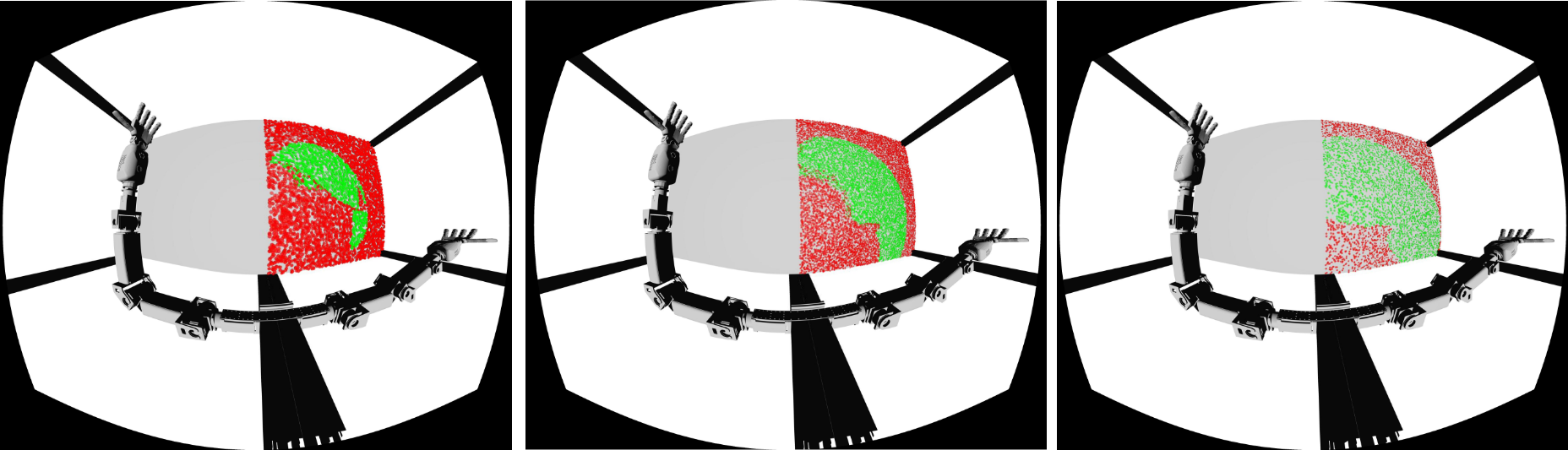}
    \caption{Left: Placement of the 10,000 collected samples (green) versus the rejected samples (red) for 6 (left), 7 (middle) and 8 (right) DoF. The images show the continuity of the IK map, i.e., to what extent a smooth IK trajectory ends up as a smooth joint trajectory if each instantaneous pose is individually passed through the IK algorithm.}
    \label{fig:distribution_all_samples_new}
\end{figure*}

Fig. \ref{fig:distribution_all_samples_new} shows the distribution of all successful grasp samples in NICOL's workspace for the three experimental conditions. Due to the rigid mechanical constraints of our defined grasping pose, the joint configuration and self-collision avoidance, the area within the defined workspace for which grasp poses can be found increases with the number of active joints. Only with eight active joints, good coverage of the workspace can be achieved. Therefore, we focus our analysis on this dataset.


\subsection{End-to-end Neurorobotic Visuomotor Learning Experiments}
\label{sec:vis_exp:neural_learning}
First, we evaluate a neural end-to-end learning approach using the neural architecture adopted from \cite{KW17}. The input to the neural network is a 96x64 pixel image from NICOL's perspective of its workspace with a single object placed in it. Depending on the dataset, the output of the \changed{network is the} seven or eight joint values of NICOL's right manipulator. The hyperparameters were optimized using Hyperopt \cite{Bergstra2013}. The model was trained with an Adam optimizer and a batch size of 35, and the MSE over the joint configuration as a loss function on a random 90-10 split of the entire dataset. Hyperparameter ranges and \changed{results are shown in} Table \ref{table:hyperparameters_cnn}, and the resulting architecture is shown in Fig. \ref{fig:neural_architecture}.

\begin{table*}[ht]
\centering
\begin{tabular}{|c|c|c|c|}
\hline
\textbf{Parameter}    & \textbf{Parameter range}  &  \textbf{Result 8 DoF} \\ \hline
\# filters conv$_1$   & 8-32 (step size 8)        & 32              \\\hline
\# filters conv$_2$   & 8-32 (step size 8)        & 8               \\\hline
kernel size conv$_1$  & 3-5 (step size 1)         & 3               \\\hline
kernel size conv$_2$  & 3-5 (step size 1)         & 5               \\\hline
\# dense\_1 neurons   & 16-1024 (step size 16)    & 256             \\\hline
\# dense\_2 neurons   & 16-256 (step size 16)     & 160             \\\hline
normalisation         & {[}0..1{]} or {[}-1..1{]} & [0..1]          \\\hline
image size            & 48x32, 96x64, 192x128     & 96x64           \\\hline
learning rate         & {[}0.01, 0.001, 0.0001, 0.00001{]} & 0.0001           \\\hline
\end{tabular}
\caption{Ranges and results of hyperparameter optimization of the end-to-end visuomotor architecture for 8 DoF.}
\label{table:hyperparameters_cnn}
\end{table*}

We use 10-fold cross-validation to obtain our results. For evaluation, we denormalize the joint output values and calculate the robotic hand's pose using the known forward kinematics of NICOL. The mean position error is 0.42 $\pm$ 0.21 meters and the mean orientation error is 7.72 $\pm$ 5.88 degrees.
Based on previous work \cite{KW17, KSSW20}, we count a grasp as successful if the position error of the resulting pose is $<$ 10 mm and the sum of orientation errors is $<$ 20 degrees. These limits are regularly exceeded. Table \ref{table:ete-results} summarizes the results with a single-digit grasp accuracy. We conclude that the end-to-end approach cannot be transferred directly to NICOL. We attribute this result to several factors: The workspace of NICOL with 100x100 cm is larger than the workspace of NICO with  30x40 cm; more importantly, the arms of NICOL afford more complex motions and more diverse joint configurations during grasping.

\begin{table*}[ht]
\centering
\begin{tabular}{|c|c|c|c|c|c|c|}
\hline
\textbf{Experiment} & \textbf{DoF} & \textbf{Mean pos. error}  & \textbf{Median pos. Error} & \textbf{Mean orient. error} & \textbf{Median orient. error} & \textbf{Grasp accuracy} \\ \hline
End-to-end          & 8   & 0.4161 $\pm$ 0.2085 m   & 0.3908 m                 & 7.72 $\pm$ 5.88 deg.  & 6.01 deg.                 & 0.01 \% \\ \hline
\end{tabular}
\caption{Results for the neural end-to-end approach for 8 DoF. Results indicate that the neural end-to-end approach cannot be transferred to the more complex kinematics of NICOL.}
\label{table:ete-results}
\end{table*}


\subsection{Modular Approach with Classical Image Processing, Neural Coordinate Transformation and IK solvers}
\label{sec:vis_exp:neurogenetic:color_based_loc}
Next, \changed{we analyze} if the low accuracy of neural end-to-end visuomotor learning can be attributed to issues in visually locating the grasp object in the image, transforming the image coordinates to world coordinates or the computation of inverse kinematics. In previous work, we address this issue using a modular, hybrid neuro-genetic approach \cite{KSSW20}. We adopt this approach to the NICOL and use three modules as shown in Fig. \ref{fig:CONDITION}: Classical image processing to locate the object in the image, a neural-network to transform the image coordinates into \changed{world coordinates, and existing state-of-the-art and novel generic IK solvers.} To extract the grasp object's position from the non-downsampled image (4208x3120) using classical image processing, we detect the top and the bottom of the grasp object with a standard color-based detector, resulting in the x, y position and the radius of the base and top parts of our grasp object. These 6 input parameters are fed into an MLP, which outputs the x- and y-coordinates of the object on the table. The network consists of one dense layer with 180 neurons and ReLU activation; the architecture is based on \cite{KSSW20} and was optimized with Hyperopt \cite{Bergstra2013}. Table \ref{table:hyperparameters_image_to_coordinate} shows the optimization ranges and results.

\begin{table}[!t]
\centering
\begin{tabular}{|c|c|c|c|}
\hline
\textbf{Parameter}  & \textbf{Parameter range}  & \textbf{Result}  \\ \hline
\# dense neurons    & 10-200 (step size 10)     & 180                    \\\hline
normalisation       & {[}0..1{]} or {[}-1..1{]} & {[}0..1{]}             \\\hline
\end{tabular}
\caption{Ranges and results of hyperparameter optimization of the neural image-to-coordinate transformation.}
\label{table:hyperparameters_image_to_coordinate}
\end{table}



We use 10-fold cross-validation for all reported results. First, we evaluate the error of the transformation. As shown in the first line in Table \ref{table:ptt-results}, the resulting mean position error is 0.0039 $\pm$ 0.0023 m, and the median position error is ~3 mm. Applying the above-established criteria for a successful grasp (obviously without considering orientation) \changed{resulted} in a grasp accuracy of 98.47\%. Next, we evaluate the complete grasp architecture by using IK solvers to compute the robot's joint configuration based on the predicted object coordinates. Table \ref{table:ptt-results} summarizes the results for the state-of-the-art IK solvers KDL, TRACK-IK and BIO-IK as well as for our novel genetic IK solver GAIKPy. Again, we apply the criteria for a successful grasp (position error is $<$ 10 mm, and the sum of orientation errors is $<$ 20 degrees). 
\changed{We achieved accuracies} of 90.52\%, 90.29\% and 89.60\% for KDL, TRAC-IK and BIO-IK respectively. In both conditions, GAIKPy shows the highest accuracy with 98.45\% and 92.85\%. 

\begin{table*}[ht]
\centering
\begin{tabular}{|c|c|c|c|c|c|c|}
\hline
\textbf{Experiment} & DoF & \textbf{Mean pos. error}  & \textbf{Median pos. Error} & \textbf{Mean orient. error} & \textbf{Median orient. error} & \textbf{Grasp accuracy} \\ \hline
transformation      & -   & 0.0039 $\pm$ 0.0023 & 0.00343      & -                             & -             & 98.47           \\\hline 
KDL                 & 8   & 0.0229 $\pm$ 0.1031 & 0.00444      & 0.519 $\pm$ 2.924 & 1.817e-05  & 90.52 \\
TRAC-IK             & 8   & 0.0244 $\pm$ 0.1068 & 0.00445      & 0.564 $\pm$ 3.042 & 1.630e-05  & 90.29 \\
BIO IK              & 8   & 0.0280 $\pm$ 0.1144 & 0.00448      & 0.686 $\pm$ 3.343 & 1.987e-04  & 89.60 \\
GAIKPy              & 8   & 0.0067 $\pm$ 0.0302 & 0.00436      & 0.337     $\pm$ 0.687     & 0.1186      & \textbf{92.85} \\ \hline
\end{tabular}
\caption{Results for the modular approach with Classical Image Processing, Neural Coordinate Transformation and IK solvers for 8 DoF. GAIKPy achieves the highest grasp accuracy of all evaluated IK solvers.}
\label{table:ptt-results}
\end{table*}

\subsection{Novel neural image-to-coordinate transformation and genetic algorithm}
\label{sec:vis_exp:neurogenetic:image_to_table}

To evaluate the limits of neural visuomotor learning, we modify the end-to-end architecture presented above to regress directly the x- and y-coordinates of the object on the table instead of the joint configurations. We then use state-of-the-art IK solvers and GAIKPy to compute a joint configuration that reaches for the output coordinate with a given hand orientation.

We use 10-fold cross-validation to obtain the results in Table \ref{table:itt-results}. Applying the criteria for a successful grasp, 
96.51\%, 96.2\% and 95.71\% for KDL, TRAC-IK and BIO-IK. 
GAIKPy shows the highest accuracy with 
99.17\%.


\begin{table*}[ht]
\centering
\begin{tabular}{|c|c|c|c|c|c|c|}
\hline
\textbf{Experiment} & DoF & \textbf{Mean pos. error}  & \textbf{Median pos. Error} & \textbf{Mean orient. error} & \textbf{Median orient. error} & \textbf{Grasp accuracy} \\ \hline
transformation      & -   & 0.0018 $\pm$ 0.0012 & 0.00158      & -                       & -             & 99.98    \\ \hline
KDL                 & 8   & 0.0191 $\pm$ 0.1034 & 0.00125      & 0.487 $\pm$ 2.834 & 1.837e-05 & 96.51 \\
TRACK-IK            & 8   & 0.0209 $\pm$ 0.1078 & 0.00125      & 0.540 $\pm$ 2.979 & 1.627e-05 & 96.20 \\
BIO-IK              & 8   & 0.0233 $\pm$ 0.1130 & 0.00126      & 0.625 $\pm$ 3.196 & 2.035e-04 & 95.71 \\
GAIKPy              & 8   & 0.0019 $\pm$ 0.0102 & 0.00122      & 0.465 $\pm$ 1.151   & 0.1841      & 99.17 \\ \hline
\end{tabular}
\caption{Results for our novel hybrid approach combining neural image-to-coordinate transformation and genetic IK with GAIKPy for 8 DoF. Again, GAIKPy achieves the highest grasp accuracy.}
\label{table:itt-results}
\end{table*}

\subsection{Discussion}
In Section \ref{sec:vis_exp:dataset}, we show that \changed{8 DoFs are needed} to cover sufficiently large parts of NICOL's intended workspace. In contrast, non-humanoid industrial robot arms often only need 6 DoF to cover their workspace. We attribute this finding to the NICOL design mimicking human motion, resulting in a more humanoid distribution of DoF along the arm and, more importantly, severe constraints on its joints. Next, we show that the increased kinematic complexity of NICOL cannot be handled by an end-to-end neural visuomotor learning approach in \changed{Section \ref{sec:vis_exp:neural_learning}}. However, we successfully apply a hybrid modular approach using image processing and a neural coordinate transformation with an IK solver \changed{in Section \ref{sec:vis_exp:neurogenetic:color_based_loc}.} We also demonstrate that NICOL integrates well with the state-of-the-art IK solvers KDL, TRACK-IK and BIO-IK. However, the best results were achieved by using our genetic algorithm-based GAIKPy IK solver.
We attribute this finding to the combination of more than 6 DoF in conjunction with joint constraints, posing a challenge for traditional state-of-the-art IK solvers, as shown in Tables \ref{table:ptt-results} and \ref{table:itt-results}. Finally, \changed{in Section \ref{sec:vis_exp:neurogenetic:image_to_table},} we positively evaluated a novel hybrid approach that uses a convolutional neural network to extract target grasp positions from an image in conjunction with an IK solver. Again, our GAIKPy approach yields the highest grasp accuracy with a success rate of over 99\%. 

This added accuracy comes at the cost of longer processing times. We give each IK solver 1 second to compute a solution. While GAIKPy fully utilizes this time, the state-of-the-art solvers \changed{show a different behavior:} In cases, where they can find a good solution, this often happens much faster, however, they don't utilize the time budget to find suitable solutions for difficult poses.

We argue that in human-robot collaboration, these slower IK solutions are less problematic. For a humanoid robot, collaborating with a human, we face different challenges compared to an industrial robotic arm. For human-robot collaboration, safe and more importantly, predictable motions are essential. For safety reasons, we need to limit the technical possible maximum joint speed of the humanoid anyway. The slower but more accurate hybrid neuro-genetic approach yields better accuracy scenarios, while still fulfilling the time constraints for collaborative human-robot scenarios.


\section{Conclusion and Outlook}
\label{sec:conclusion}

\changed{
We present NICOL, a novel semi-humanoid robotic platform designed for research in social robotics and physical human-robot collaboration. NICOL, combining the well-received social interaction features of the NICO platform\cite{KSMNHW17} with adult-sized manipulators, fills a gap in the current state of the art. It offers an affordable and open platform for research involving robots collaborating with and learning from humans in scenarios demanding advanced object manipulation and interaction abilities.
}

\changed{
Our evaluation of NICOL primarily focuses on its inverse kinematics capability in the context of a reach-for-grasp task, which is an essential skill for humanoid robots. We positively evaluate our main design goal for NICOL as a successor of the child-sized NICO, and to create a robotic platform that has a workspace and can handle objects comparable to an adult human, thus being able to be used in physical collaboration scenarios. We also address the issue of standard IK solvers having problems with kinematic chains that have more than 6 DoF, while at the same time having strict constraints on the individual joint limits to mimic the human range of motion. We evaluate neural end-to-end learning approaches for visuomotor learning and compare state-of-the-art IK solvers against a novel hybrid neuro-genetic IK approach. We show that neural end-to-end learning is challenging due to the more complex kinematics of NICOL, and overcome this challenge with a neuro-genetic approach. Furthermore, we demonstrate that our novel hybrid neuro-genetic approach outperforms classical IK solvers in terms of accuracy by taking advantage of the relaxed time constraints in human-robot collaboration scenarios.
}

\changed{In future work, we will thoroughly evaluate NICOL's abilities in different collaborative real-world tasks where it will learn from and assist human participants. One such task is assembly, where NICOL will be asked to pick up and hand over different components to a human participant. Besides the motoric challenge, such tasks required shared attention on objects, social cues regarding the handover and verbal communication to coordinate efforts. A second research scenario is learning from a human through observations of actions. To this end, we will design neurocognitive architectures that can segment perceived demonstration into meaningful units and relate them in a semantic way to objects on the table so that NICOL can perform the demonstrated actions from its perspective, even in a changed environment, e.g., with objects placed on the table differently. Both scenarios combine NICOL's motor abilities with social interaction capability. Furthermore, we will use NICOL as a platform for ongoing research for neural and neuro-hybrid IK and motion planning approaches for robots with a restricted, human-like range of motion.}


\bibliographystyle{unsrt}
\bibliography{NICOL}

\begin{thebibliography}{10}

\bibitem{Thomaz:2006}
A.~L. {Thomaz}, G.~{Hoffman}, and C.~{Breazeal}.
\newblock Reinforcement learning with human teachers: Understanding how people
  want to teach robots.
\newblock In {\em ROMAN 2006 - The 15th IEEE International Symposium on Robot
  and Human Interactive Communication}, pages 352--357, Sep. 2006.

\bibitem{Srinivasan:2016}
Vasant Srinivasan and Leila Takayama.
\newblock Help me please: Robot politeness strategies for soliciting help from
  humans.
\newblock In {\em Proceedings of the 2016 CHI conference on human factors in
  computing systems}, pages 4945--4955. ACM, 2016.

\bibitem{KPSHW20}
Matthias Kerzel, Theresa Pekarek-Rosin, Erik Strahl, Stefan Heinrich, and
  Stefan Wermter.
\newblock Teaching {NICO} how to grasp: An empirical study on crossmodal social
  interaction as a key factor for robots learning from humans.
\newblock {\em Frontiers in Neurorobotics}, Jun 2020.

\bibitem{mutlu2013coordination}
Bilge Mutlu, Allison Terrell, and Chien-Ming Huang.
\newblock Coordination mechanisms in human-robot collaboration.
\newblock In {\em Proceedings of the Workshop on Collaborative Manipulation,
  8th ACM/IEEE International Conference on Human-Robot Interaction}, pages
  1--6. Citeseer, 2013.

\bibitem{KSMNHW17}
Matthias Kerzel, Erik Strahl, Sven Magg, Nicol\'as Navarro-Guerrero, Stefan
  Heinrich, and Stefan Wermter.
\newblock {NICO} -- {N}euro-{I}nspired {CO}mpanion: {A} developmental humanoid
  robot platform for multimodal interaction.
\newblock In {\em Proceedings of the IEEE International Symposium on Robot and
  Human Interactive Communication (RO-MAN)}, pages 113--120, Aug 2017.

\bibitem{mivseikis2018transfer}
Justinas Mi{\v{s}}eikis, Inka Brijacak, Saeed Yahyanejad, Kyrre Glette,
  Ole~Jakob Elle, and Jim Torresen.
\newblock Transfer learning for unseen robot detection and joint estimation on
  a multi-objective convolutional neural network.
\newblock In {\em 2018 IEEE International Conference on Intelligence and Safety
  for Robotics (ISR)}, pages 337--342. IEEE, 2018.

\bibitem{UR5}
{Universal Robots}.
\newblock {UR5 Technical Specifications}.
\newblock Accessed 2020-12-27.[Online]. Available at:
  http://www.universal-robots.com.

\bibitem{bischoff2010kuka}
Rainer Bischoff, Johannes Kurth, G{\"u}nter Schreiber, Ralf Koeppe, Alin
  Albu-Sch{\"a}ffer, Alexander Beyer, Oliver Eiberger, Sami Haddadin, Andreas
  Stemmer, Gerhard Grunwald, et~al.
\newblock The {KUKA-DLR} lightweight robot arm-a new reference platform for
  robotics research and manufacturing.
\newblock In {\em ISR 2010 (41st international symposium on robotics) and
  ROBOTIK 2010 (6th German conference on robotics)}, pages 1--8. VDE, 2010.

\bibitem{pandarobot}
Franka Emika.
\newblock Franka research 3.
\newblock \url{https://www.franka.de/research}, 2022.
\newblock Accessed 2022-12-27.

\bibitem{metta2010icub}
Giorgio Metta, Lorenzo Natale, Francesco Nori, Giulio Sandini, David Vernon,
  Luciano Fadiga, Claes Von~Hofsten, Kerstin Rosander, Manuel Lopes, Jos{\'e}
  Santos-Victor, et~al.
\newblock The {iCub} humanoid robot: An open-systems platform for research in
  cognitive development.
\newblock {\em Neural Networks}, 23(8-9):1125--1134, 2010.

\bibitem{al2012furhat}
Samer Al~Moubayed, Jonas Beskow, Gabriel Skantze, and Bj{\"o}rn Granstr{\"o}m.
\newblock Furhat: {A} back-projected human-like robot head for multiparty
  human-machine interaction.
\newblock In {\em Cognitive Behavioural Systems}, pages 114--130. Springer,
  2012.

\bibitem{ha2011development}
Inyong Ha, Yusuke Tamura, Hajime Asama, Jeakweon Han, and Dennis~W Hong.
\newblock Development of open humanoid platform {DARwIn-OP}.
\newblock In {\em SICE Annual Conference 2011}, pages 2178--2181. IEEE, 2011.

\bibitem{schwarz2013humanoid}
Max Schwarz, Julio Pastrana, Philipp Allgeuer, Michael Schreiber, Sebastian
  Schueller, Marcell Missura, and Sven Behnke.
\newblock Humanoid {TeenSize} open platform {NimbRo-OP}.
\newblock In {\em Robot Soccer World Cup}, pages 568--575. Springer, 2013.

\bibitem{AllgeuerIguhop}
Philipp Allgeuer, Hafez Farazi, Michael Schreiber, and Sven Behnke.
\newblock Child-sized {3D} printed igus humanoid open platform.
\newblock In {\em 15th International Conference on Humanoid Robots
  (Humanoids)}, Seoul, Korea, 2015.

\bibitem{FichtNop2X}
Grzegorz Ficht, Hafez Farazi, Andr\'e Brandenburger, Diego Rodriguez, Dmytro
  Pavlichenko, Philipp Allgeuer, Mojtaba Hosseini, and Sven Behnke.
\newblock {NimbRo-OP2X}: {A}dult-sized open-source {3D} printed humanoid robot.
\newblock In {\em 18th International Conference on Humanoid Robots
  (Humanoids)}, Beijing, China, 2018.

\bibitem{Pandey:2018:pepper}
Amit~Kumar Pandey and Rodolphe Gelin.
\newblock A mass-produced sociable humanoid robot: Pepper: The first machine of
  its kind.
\newblock {\em IEEE Robotics \& Automation Magazine}, PP:1--1, 07 2018.

\bibitem{Lapeyre2014_Poppy}
Matthieu Lapeyre, Pierre Rouanet, Jonathan Grizou, Steve N'Guyen, Alexandre
  Le~Falher, Fabien Depraetre, and Pierre-Yves Oudeyer.
\newblock Poppy: {O}pen source {3D printed} robot for experiments in
  developmental robotics.
\newblock In {\em International Conference on Development and Learning and on
  Epigenetic Robotics (ICDL-EpiRob)}, pages 173--174, 2014.

\bibitem{stasse2017talos}
Olivier Stasse, Thomas Flayols, Rohan Budhiraja, Kevin Giraud-Esclasse, Justin
  Carpentier, Joseph Mirabel, Andrea Del~Prete, Philippe Sou{\`e}res, Nicolas
  Mansard, Florent Lamiraux, et~al.
\newblock {TALOS}: A new humanoid research platform targeted for industrial
  applications.
\newblock In {\em 2017 IEEE-RAS 17th International Conference on Humanoid
  Robotics (Humanoids)}, pages 689--695. IEEE, 2017.

\bibitem{asfour2019armar}
Tamim Asfour, Mirko Waechter, Lukas Kaul, Samuel Rader, Pascal Weiner, Simon
  Ottenhaus, Raphael Grimm, You Zhou, Markus Grotz, and Fabian Paus.
\newblock {ARMAR-6}: A high-performance humanoid for human-robot collaboration
  in real-world scenarios.
\newblock {\em IEEE Robotics \& Automation Magazine}, 26(4):108--121, 2019.

\bibitem{dean2017tomm}
Emmanuel Dean-Leon, Brennand Pierce, Florian Bergner, Philipp Mittendorfer,
  Karinne Ramirez-Amaro, Wolfgang Burger, and Gordon Cheng.
\newblock {TOMM}: Tactile omnidirectional mobile manipulator.
\newblock In {\em 2017 IEEE International Conference on Robotics and Automation
  (ICRA)}, pages 2441--2447. IEEE, 2017.

\bibitem{fetch}
Fetch Robotics.
\newblock Robotics automation for warehousing, {3PLs}, distribution,
  manufacturing, 2019.

\bibitem{parmiggiani2017designR1}
Alberto Parmiggiani, Luca Fiorio, Alessandro Scalzo, Anand~Vazhapilli
  Sureshbabu, Marco Randazzo, Marco Maggiali, Ugo Pattacini, Hagen Lehmann,
  Vadim Tikhanoff, Daniele Domenichelli, et~al.
\newblock The design and validation of the {R1} personal humanoid.
\newblock In {\em 2017 IEEE/RSJ International Conference on Intelligent Robots
  and Systems (IROS)}, pages 674--680. IEEE, 2017.

\bibitem{CKSBW17}
Nikhil Churamani, Matthias Kerzel, Erik Strahl, Pablo Barros, and Stefan
  Wermter.
\newblock Teaching emotion expressions to a human companion robot using deep
  neural architectures.
\newblock In {\em International Joint Conference on Neural Networks (IJCNN)},
  pages 627--634, Anchorage, Alaska, May 2017. IEEE.

\bibitem{Bartneck:2009}
Christoph Bartneck, Dana Kuli{\'{c}}, Elizabeth Croft, and Susana Zoghbi.
\newblock Measurement instruments for the anthropomorphism, animacy,
  likeability, perceived intelligence, and perceived safety of robots.
\newblock {\em International Journal of Social Robotics}, 1(1):71--81, Jan
  2009.

\bibitem{CBSW18}
Nikhil Churamani, Pablo Barros, Erik Strahl, and Stefan Wermter.
\newblock Learning empathy-driven emotion expressions using affective
  modulations.
\newblock In {\em Proceedings of the International Joint Conference on Neural
  Networks (IJCNN 2018)}, pages 1400--1407. IEEE, IEEE, Jul 2018.

\bibitem{NABCFHMMNNSSGHNSTWW17}
Hwei~Geok Ng, Paul Anton, Marc Br{\"u}gger, Nikhil Churamani, Erik
  Fließwasser, Thomas Hummel, Julius Mayer, Waleed Mustafa, Thi Linh~Chi
  Nguyen, Quan Nguyen, Marcus Soll, Sebastian Springenberg, Sascha Griffiths,
  Stefan Heinrich, Nicol\'as Navarro-Guerrero, Erik Strahl, Johannes Twiefel,
  Cornelius Weber, and Stefan Wermter.
\newblock Hey robot, why don't you talk to me?
\newblock In {\em Proceedings of the IEEE International Symposium on Robot and
  Human Interactive Communication (RO-MAN)}, pages 728--731, Aug 2017.

\bibitem{BXABCNNSAGHSWW19}
Hadi Beik-Mohammadi, Nikoletta Xirakia, Fares Abawi, Irina Barykina, Krishnan
  Chandran, Gitanjali Nair, Cuong Nguyen, Daniel Speck, Tayfun Alpay, Sascha
  Griffiths, Stefan Heinrich, Erik Strahl, Cornelius Weber, and Stefan Wermter.
\newblock Designing a personality-driven robot for a human-robot interaction
  scenario.
\newblock In {\em 2019 IEEE International Conference on Robotics and Automation
  (ICRA)}, pages 4317--4324, Montreal, Canada, May 2019.

\bibitem{mori2012uncanny}
Masahiro Mori, Karl~F MacDorman, and Norri Kageki.
\newblock {The Uncanny Valley [From the Field]}.
\newblock {\em IEEE Robotics \& Automation Magazine}, 19(2):98--100, 2012.

\bibitem{coleman2014reducing}
David Coleman, Ioan Sucan, Sachin Chitta, and Nikolaus Correll.
\newblock Reducing the barrier to entry of complex robotic software: {A}
  {MoveIt!} case study.
\newblock {\em Journal of Software Engineering for Robotics}, 5(1):3--16, 2014.

\bibitem{kdl-url}
R.~Smits.
\newblock {KDL}: {K}inematics and {D}ynamics {L}ibrary.
\newblock \url{http://www.orocos.org/kdl}.

\bibitem{tracikBeeson}
Patrick Beeson and Barrett Ames.
\newblock {TRAC-IK}: {A}n open-source library for improved solving of generic
  inverse kinematics.
\newblock In {\em 2015 IEEE-RAS 15th International Conference on Humanoid
  Robots (Humanoids)}, pages 928--935, 2015.

\bibitem{bioikStarke}
Sebastian Starke, Norman Hendrich, and Jianwei Zhang.
\newblock A memetic evolutionary algorithm for real-time articulated kinematic
  motion.
\newblock In {\em 2017 IEEE Congress on Evolutionary Computation (CEC)}, pages
  2473--2479, 2017.

\bibitem{bioikmoveitRuppel}
Philipp~Sebastian Ruppel.
\newblock Performance optimization and implementation of evolutionary inverse
  kinematics in {ROS}.
\newblock Master's thesis, University of Hamburg, Department of Computer
  Science, 2017.

\bibitem{bioikserviceGithub}
{BioIK Service}: {ROS} service for calling {BioIK} from {P}ython or {J}ava.
\newblock \url{https://github.com/TAMS-Group/bio_ik_service}.
\newblock Accessed: 2023-10-06.

\bibitem{KW17}
Matthias Kerzel and Stefan Wermter.
\newblock Neural end-to-end self-learning of visuomotor skills by environment
  interaction.
\newblock In {\em Artificial Neural Networks and Machine Learning – ICANN
  2017}, volume 10613 of {\em Lecture Notes in Computer Science}, pages 27--34.
  Springer, Cham, Sep 2017.

\bibitem{marslandmachine}
Stephen Marsland.
\newblock {\em Machine Learning: An Algorithmic Perspective, Second Edition}.
\newblock Chapman \& Hall/CRC, 2nd edition, 2014.

\bibitem{KSSW20}
Matthias Kerzel, Josua Spisak, Erik Strahl, and Stefan Wermter.
\newblock Neuro-genetic visuomotor architecture for robotic grasping.
\newblock In {\em Artificial Neural Networks and Machine Learning – ICANN
  2020}, LNCS, pages 533--545. Springer, 2020.

\bibitem{Bergstra2013}
James Bergstra, Daniel Yamins, and David Cox.
\newblock Making a science of model search: Hyperparameter optimization in
  hundreds of dimensions for vision architectures.
\newblock In {\em 30th International Conference on Machine Learning (ICML
  2013)}, pages 115--123, 2013.

\end{thebibliography}

\phantomsection

\ifarxiv\else

\begin{IEEEbiography}[{\includegraphics[width=1in,height=1.25in,clip,keepaspectratio]{bios/kerzel.jpg}}]{Matthias Kerzel}
received his MSc and PhD in computer science from the
University of Hamburg, Germany. He is currently a postdoctoral research
and teaching associate at the Knowledge Technology Group of Prof. Stefan
Wermter at the University of Hamburg. He has given lectures on Knowledge
Processing in Intelligent Systems, Neural Networks and Bio-inspired
Artificial Intelligence. He is currently the Secretary of the European Neural Network Society (ENNS) and
worked in the organizing committee of the ICANN conferences. His
research interests are in developmental neurorobotics, hybrid
neurosymbolic architectures, explainable AI and human-robot interaction.
He is currently involved in the international SFB/TRR-169 large-scale
project on crossmodal learning.
\end{IEEEbiography}

\begin{IEEEbiography}[{\includegraphics[width=1in,height=1.25in,clip,keepaspectratio]{bios/allgeuer.jpg}}]{Philipp Allgeuer}
received his BEng/BSc with 1st class Honours from the University of Adelaide, Australia, graduating with the overall top University Medal for that year. He received his PhD in computer science from the University of Bonn in the field of humanoid robotics and bipedal locomotion, and is currently a postdoctoral research associate at the Knowledge Technology Group at the University of Hamburg. His research focus lies in making advances in deep learning methods and architectures, and uniting these with physical robot platforms to provide better human-robot interaction experiences and capabilities.
\end{IEEEbiography}

\begin{IEEEbiography}[{\includegraphics[width=1in,height=1.25in,clip,keepaspectratio]{bios/strahl.jpg}}]{Erik Strahl}
is the technical engineer in the Knowledge Technology Institute in the Department of Informatics. He studied computer science and law at the University of Applied Science in Hamburg and at the University of Hamburg. One focus of his work is the development and preparation of robots for the research of the Knowledge Technology Institute on the software and hardware side. In this position, he participated in the development of the NICO and NICOL robots and the usage of them in many research projects.
\end{IEEEbiography}

\begin{IEEEbiography}[{\includegraphics[width=1in,height=1.25in,clip,keepaspectratio]{bios/frick.jpg}}]{Nicolas Frick}
completed vocational training and worked in the industrial sector as a Mechatronic engineer before moving to academia. He received his BSc in applied computer science from the University of Duisburg-Essen in the field of embedded systems. He is currently finishing his MSc in computer science at the University of Hamburg, and is part of the NICOL humanoid robot engineering team. In his research, he focuses mainly on the applicability of deep learning methods to humanoid robotic grasping.
\end{IEEEbiography}

\begin{IEEEbiography}[{\includegraphics[width=1in,height=1.25in,clip,keepaspectratio]{bios/habekost.jpg}}]{Jan-Gerrit Habekost}
received his BSc in business informatics from the Carl von Ossietzky University Oldenburg and is currently finishing his MSc in computer science at the University of Hamburg. He is a student assistant at the Knowledge Technology Group at the University of Hamburg. His research focuses on humanoid robotics, deep learning and neuro-robotic applications.
\end{IEEEbiography}

\begin{IEEEbiography}[{\includegraphics[width=1in,height=1.25in,clip,keepaspectratio]{bios/eppe.jpg}}]{Manfred Eppe}
is Chief Engineer and Head of the Robotics Lab of the Institute for Data Science Foundations at the Hamburg University of Technology. His current research is in cognitive robotics and machine learning. Previously, he was a postdoctoral research associate at the Knowledge Technology Institute at the University of Hamburg. He has also held positions at the University of California at Berkeley, the Artificial Intelligence Research Institute in Barcelona, and the University of Bremen. 
Dr. Eppe has published in numerous renowned venues, including Nature Machine Intelligence, Artificial Intelligence, and the International Conference on Intelligent Robots and Systems (IROS). 
He has coordinated several research projects funded by the German Research Foundation (DFG) and the Volkswagen Stiftung. His research focus lies in computational cognitive models of representation learning and reinforcement learning for simulated and physical agents and robots.
\end{IEEEbiography}

\begin{IEEEbiography}[{\includegraphics[width=1in,height=1.25in,clip,keepaspectratio]{bios/wermter.jpg}}]{Stefan Wermter}
is a Full Professor at the University of Hamburg, Germany, and Director of the Knowledge Technology Institute in the Dept. of Informatics. He has previously held positions at the University of Dortmund, University of Massachusetts, the International Computer Science Institute in Berkeley, and the University of Sunderland. His main research interests are in neural networks, hybrid knowledge technology, neuroscience-inspired computing, cognitive robotics, natural language processing and human-robot interaction. He has been associate editor of the journal `IEEE Transactions on Neural Networks and Learning Systems' and he is on the advisory board of `Connection Science' and `International Journal for Hybrid Intelligent Systems', and he is on the editorial board of the journals `Cognitive Computation' and `Journal of Computational Intelligence'. He is the coordinator of the international doctoral training network TRAIL, co-coordinator of the international collaborative research center on Crossmodal Learning (TRR-169), and he is currently serving as the elected President of the European Neural Network Society.
\end{IEEEbiography}

\fi

\EOD

\end{document}